\documentclass{uai2020}

\author{
  T\'arik S.~Salem \\
  \Email{tariks@alumni.ntnu.no} \\
  \And
  Helge Langseth \\
  \Email{helgel@ntnu.no} \\
  \And
  Heri Ramampiaro \\
  \Email{heri@ntnu.no} \\
  \Affiliation
  Norwegian University of Science and Technology (NTNU)
}

\begin{document}

\maketitle

\begin{abstract}
Prediction intervals are a machine- and human-interpretable way to represent
predictive uncertainty in a regression analysis. In this paper, we present a
method for generating prediction intervals along with point estimates from an
ensemble of neural networks.
We propose a multi-objective loss function fusing quality measures related to
prediction intervals and point estimates, and a penalty function, which
enforces semantic integrity of the results and stabilizes the training
process of the neural networks.
The ensembled prediction intervals are aggregated as a split normal mixture
accounting for possible multimodality and asymmetricity of the posterior
predictive distribution, and resulting in prediction intervals that capture
aleatoric and epistemic uncertainty.
Our results show that both our quality-driven loss function and our aggregation
method contribute to well-calibrated prediction intervals and point estimates.
\end{abstract}

\section{INTRODUCTION}
\label{sec:introduction}

Quantifying predictive uncertainty of machine learning models is crucial in applications, e.g., mission-critical systems, where it is essential to know when the model is not able to provide accurate predictions.
In decision support systems, providing the human with an additional information
about the uncertainty of a prediction may decrease the response time and
increase the accuracy of an action. It can also positively contribute in
building trust and understanding towards the machine learning system and its correct facilitation.
In this work, we focus on the quantification of predictive uncertainty for
the regression task, specifically in the form of prediction intervals which have
probabilistic interpretation and which are interpretable for both humans and
machines.

Prediction interval (PI) is an estimate representing predictive uncertainty in
the form of two values between which a future observation will fall with a
certain probability.
Well-calibrated/high-quality prediction intervals are as narrow as possible
while attaining the desired coverage probability.

\begin{figure}[t]
  \vskip 0.2in
  \begin{center}
  \centerline{\includegraphics[width=\columnwidth]{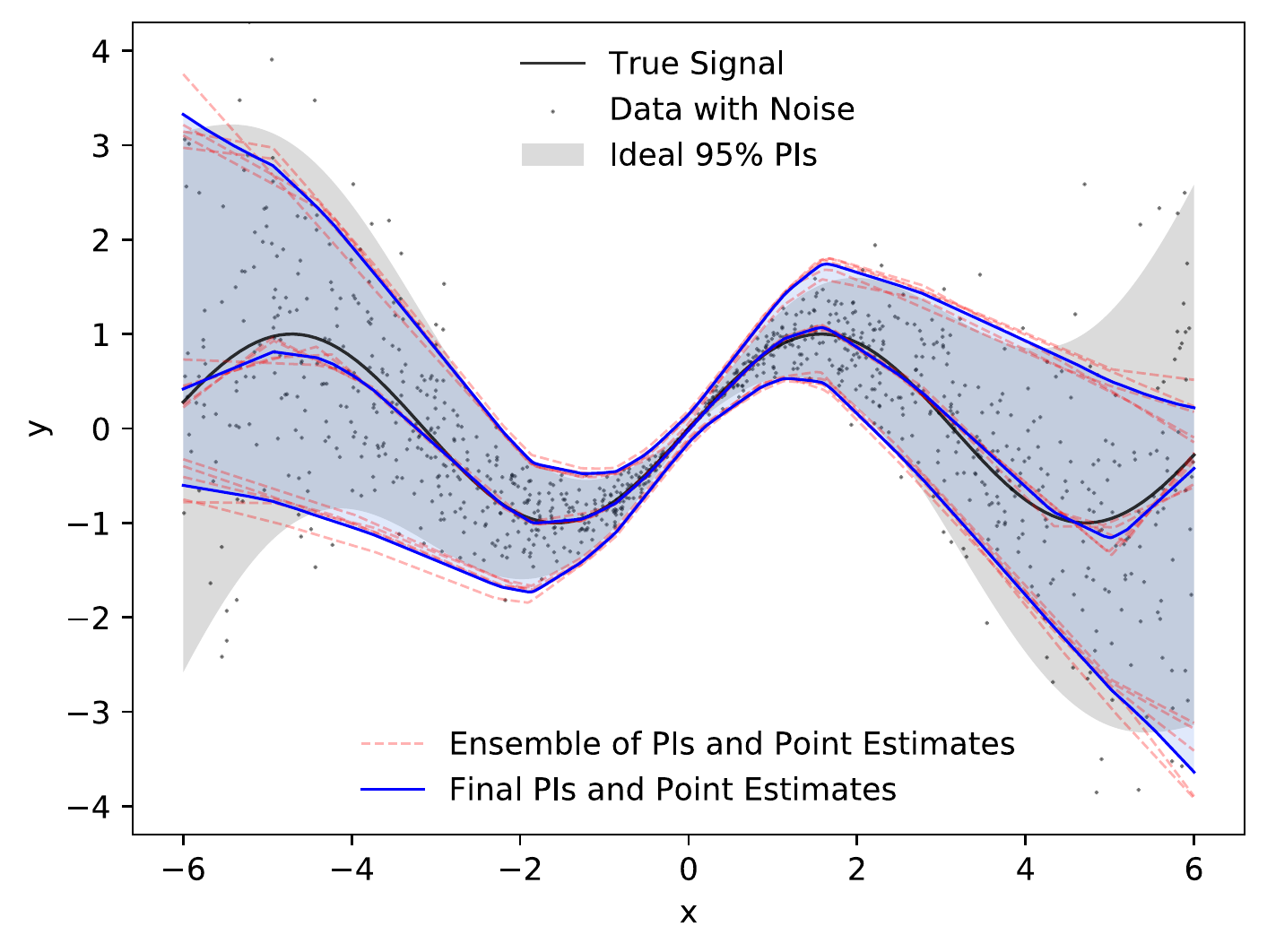}}
  \caption{A toy example demonstrating our method for generating PIs from a split normal mixture of quality-driven deep ensembles. The final PIs are accounting for aleatoric and epistemic uncertainty. The synthetic dataset is a sinusoid with Gaussian noise and sparsified samples. The magnitude of Gaussian noise and sparsification is increasing with distance from the center ($x=0$).}
  \label{fig:toy-example}
  \end{center}
  \vskip -0.2in
\end{figure}

While neural networks (NNs) are powerful function approximators, they are poor at representing
predictive uncertainty. Previously, methods adopting a Bayesian approach in the
context of neural networks \citetext{e.g., \citealp{graves:2011, blundell:2015,
hernandez:2015, krueger:2017, louizos:2017, pawlowski:2017, wu:2019, izmailov:2019}} represented the
state-of-the-art for providing predictive uncertainty estimates. See \citeauthor{yao:2019} \yrcite{yao:2019} for an overview. Recently, a number of non-Bayesian yet probabilistic approaches provide competitive predictive
uncertainty estimates~\cite{lakshminarayanan:2017, pearce:2018,
tagasovska:2019}. In this paper, we focus on and extend the latter branch of research.

Our work builds on the findings of \citeauthor{pearce:2018}
\citeyearpar{pearce:2018} which is inspired by works of
\citeauthor{khosravi:2011} \citeyearpar{khosravi:2011} and
\citeauthor{lakshminarayanan:2017} \citeyearpar{lakshminarayanan:2017}.
\citeauthor{pearce:2018} \citeyearpar{pearce:2018} presented a method based on
aggregating ensembled PIs from a set of NNs optimized with respect to a so-called
quality-driven loss function (see Section~\ref{sec:original-quality-driven-ensembles}).

Although the aforementioned work has achieved promising results of well-calibrated PIs, it has several limitations that need to be addressed. In our continuation, we primarily address three main limitations of the state-of-the-art method: (1) The inability to generate point estimates, (2) the theoretically weakly justified method for aggregation of ensembled PIs, and (3) the fragile training process. With this in mind, the main contributions of this work can be summarized as follows:

\begin{itemize}
  \item{
  We retrofit the quality-driven loss function:
  \begin{itemize}
    \item{
    With a point estimate loss (particularly MSE or mean  squared error) to be
    able to draw point estimates and PIs from the same generative distribution.
    }
    \item{
    With a penalty function adding a constraint and thus enforcing the integrity of
    the results, i.e. avoiding crossing of PI boundaries or point estimates out
    of PI bounds, and consequently stabilizing/strengthening the training process.
    }
  \end{itemize}
  }
  \item{
  We propose a new aggregation method for ensembles of PIs with
  point estimates. It is fitting a split normal mixture~\cite{wallis:2014} providing tighter and theoretically well-founded aggregates of PIs.
  }
  \item{
  We propose an analytical approach to parameter initialization for the
  parameter fitting process of a split normal probability density function,
  thus increasing the success of fitting and accelerating it compared to
  random initialization.
  }
\end{itemize}

In addition to the above, we provide important insights and guidelines for hyper-parameter
search contributing to reproducibility and reliable model fitting, and we
suggest directions for future research.

\section{BACKGROUND}
\label{sec:background}

In this section, we formally introduce the predictive uncertainty and
prediction intervals, we consider methods for quantifying predictive uncertainty
in NNs, especially in the form of prediction intervals, and we provide insights into the work of \citeauthor{pearce:2018} \yrcite{pearce:2018} which we build on.

\subsection{Predictive Uncertainty}
\label{sec:predictive-uncertainty}

The sources of predictive uncertainty can be categorized into aleatoric and
epistemic uncertainty~\cite{kiureghian:2009}.

The aleatoric or aleatory or data uncertainty is also known as the irreducible
uncertainty, i.e. it can not be reduced either through model or data. It arises
from an inherent and irreducible data noise. The aleatoric uncertainty can be
captured by learning the conditional distribution between the target and the input variables.
The aleatoric uncertainty can further be characterized as homoskedastic (homoscedastic) if the irreducible noise is constant across random variables or as heteroskedastic (heteroscedastic) if contrary.

In contrast to aleatoric uncertainty, the epistemic uncertainty is reducible, and it can be further decomposed into model uncertainty and distributional uncertainty.
Model uncertainty can be caused by model bias or parameter uncertainty due to
insufficient data. Distributional uncertainty can be caused by the mismatch
between training and test set~\cite{malinin:2018}, and it is often described
as a part of the model uncertainty.

Generally, the aim is to reduce epistemic uncertainties. However, constraints
emerging from model or data usually do not allow it.

Please note that predictive uncertainty is a broad concept, and the literature is
inconsistent in the terminology. We follow the taxonomy found in
\cite{kiureghian:2009, pearce:2018, malinin:2018}.

\subsection{Prediction Intervals}
\label{sec:prediction-intervals}

Given an input $\mathbf{x}^{(i)}$, a prediction interval $[\hat{y}_{L}^{(i)}, \hat{y}_{U}^{(i)}]$ of a sample $i$ captures the future observation (target variable) $y^{(i)}$ with the probability equal or greater than $\gamma \in [0, 1]$ (eq.~\ref{eq:prediction-interval}). The value of $\gamma$ is commonly set to $0.95$ or $0.99$. Common in the literature is an alternative notation with $\alpha$.
\begin{equation}
  \label{eq:prediction-interval}
  \operatorname{Pr}\left(\hat{y}_{L}^{(i)} \leq y^{(i)} \leq \hat{y}_{U}^{(i)}\right) \geq \gamma = (1-\alpha)
\end{equation}
Given $n$ samples, the quality of the generated prediction intervals is assessed
by measuring the prediction interval coverage probability (PICP)
\begin{equation*}
  \label{eq:picp}
  \mathit{PICP} = \frac{c}{n}
\end{equation*}
where
\begin{equation}
  \label{eq:c}
  c = \sum_{i=1}^{n} k_i
\end{equation}
for
\begin{equation}
  \label{eq:k_i}
  k_i =
  \begin{cases}
    1 & \text{if}\ \hat{y}_{L}^{(i)} \leq y^{(i)} \leq \hat{y}_{U}^{(i)}, \\
    0 & \text{otherwise},
  \end{cases}
\end{equation}
and by measuring the mean prediction interval width (MPIW)
\begin{equation*}
  \label{eq:mpiw}
  \mathit{MPIW} = \frac{1}{n}\sum_{i=1}^{n} \hat{y}_{U}^{(i)} - \hat{y}_{L}^{(i)}
\end{equation*}
or its normalized version NMPIW
\begin{equation}
  \label{eq:nmpiw}
  \mathit{NMPIW} = \frac{\mathit{MPIW}}{r}
\end{equation}
where $r = max(\vectorsym{y}) - min(\vectorsym{y})$. Please note that MPIW assumes $\hat{y}_{U}^{(i)} \geq \hat{y}_{L}^{(i)}$, i.e. no crossing of PI bounds. Ignoring this fact can lead to misleading results.

It is desired to achieve $\mathit{PICP} \geq \gamma$ while having MPIW as small
as possible.

\subsection{Related Work}
\label{sec:related-work}

Following is a concise exploration of methods based on NNs capable of quantifying predictive uncertainty in a regression analysis. Methods adopting the Bayesian
approach (such as BNN or Bayesian neural networks) are a prominent group \citetext{\citealp{graves:2011, blundell:2015, hernandez:2015, krueger:2017,
louizos:2017, pawlowski:2017, wu:2019, izmailov:2019}; etc.} and represented for long the
state-of-the-art to estimate predictive uncertainty. We explore alternative
non-Bayesian yet probabilistic methods representing competitive and possibly a
state-of-the-art alternative.

Interpreting Monte Carlo dropout (MC-dropout) at test time as approximate
Bayesian inference~\cite{gal:2016} has been a widely used method to quantify
predictive uncertainty, mainly due to its scalability and simplicity.
Interpretation of MC-dropout as an ensemble model combination motivated
consecutive research on ensemble-based methods which have been shown to be
superior in generating predictive uncertainty to MC-dropout or even
Bayesian methods~\cite{lakshminarayanan:2017}. Recently, it has been shown that
deep ensembles with random initialization may explore different modes in
function space, and therefore perform well in exploring model uncertainty. That
is in contrast to subspace sampling methods (e.g. MC-dropout, weight
averaging) which may generate a set of diverse functions but still in the vicinity of
the starting point and thus generating insufficiently diverse predictions
\cite{fort:2019}. Also recently, the robustness of uncertainty quantification
methods under dataset shift was investigated, and although all compared methods
have been deceptible to increasing dataset shifts, deep ensembles have shown the
greatest robustness~\cite{ovadia:2019}.

More recently, an alternative line of work employs a loss function to
optimize for generating specifically prediction intervals.
A method based on quantile regression is utilizing the so-called pinball
loss function~\cite{koenker:2005} to optimize a single NN for the desired PIs
and so-called Orthonormal Certificates to account for epistemic uncertainty
\cite{tagasovska:2019}.
A method called LUBE~\cite{khosravi:2011} constructs prediction intervals using
quality-driven loss function utilizing quality metrics PICP and NMPIW. LUBE
applies simulated annealing for training a NN with respect to the quality-driven
loss function. Inspired by \citeauthor{khosravi:2011} \yrcite{khosravi:2011} and
\citeauthor{lakshminarayanan:2017} \yrcite{lakshminarayanan:2017}, a slightly
modified quality-driven loss function optimized for gradient descent was proposed
\cite{pearce:2018} and extended with ensembles to account for epistemic
uncertainty. Although presenting favorable results, both methods deliver only
prediction intervals without point estimates.

\subsection[Quality-Driven Ensembles by Pearce et al. (2018)]{Quality-Driven Ensembles by \mbox{\citeauthor{pearce:2018} \yrcite{pearce:2018}}}
\label{sec:original-quality-driven-ensembles}

Our work builds on~\cite{pearce:2018} henceforth referred to as the original
quality-driven ensembles or shortly original QDE (in results annotated as SEM-QD). First, the quality-driven loss function is described (annotated as QD). Second, the aggregation method is described (annotated as SEM).

\subsubsection{Loss Function}
\label{sec:original-loss-function}

Parameters of each NN in an ensemble of size $m$ are optimized with respect to
the loss function $\mathcal{L}_{QD}$ (eq.~\ref{eq:original-qde}). The loss
function operates on mini-batches of size $n$.
\begin{equation}
  \label{eq:original-qde}
  \mathcal{L}_{QD} = \mathcal{L}_{MPIW} + \lambda \frac{n}{\alpha (1 - \alpha)} \mathcal{L}_{PICP}
\end{equation}
$\mathcal{L}_{MPIW}$ defined in eq.\ (\ref{eq:loss-mpiw}) is optimizing the width of the PIs
capturing an observation. Due to variables $c$ and $k_i$ from eq.\ (\ref{eq:c})
and (\ref{eq:k_i}), $\mathcal{L}_{MPIW}$ only includes those samples for which the observation is inside the PI.
\begin{equation}
  \label{eq:loss-mpiw}
  \begin{aligned}
  \mathcal{L}_{MPIW} = \frac{1}{c}\sum_{i=1}^{n} (\hat{y}_{U}^{(i)} - \hat{y}_{L}^{(i)}) \cdot k_i
 \end{aligned}
\end{equation}
$\mathcal{L}_{PICP}$ defined in eq.\ (\ref{eq:loss-picp}) is optimizing the coverage probability of the PIs penalizing only if the PICP is below the desired $\gamma$.
\begin{equation}
  \label{eq:loss-picp}
  \begin{aligned}
  \mathcal{L}_{PICP} = \max (0, (1 - \alpha) - PICP)^2
 \end{aligned}
\end{equation}
Expanded, we get
\begin{equation*}
  \label{eq:original-qde-expanded}
  \begin{aligned}
  \mathcal{L}_{QD} ={} & \frac{1}{c}\sum_{i=1}^{n} (\hat{y}_{U}^{(i)} - \hat{y}_{L}^{(i)}) \cdot k_i \\
  & + \lambda \frac{n}{\alpha (1 - \alpha)} \max (0, (1 - \alpha) - PICP)^2.
  \end{aligned}
\end{equation*}
The hyper-parameter (Lagrangian) $\lambda$ controls the importance of
$\mathcal{L}_{PICP}$ with respect to $\mathcal{L}_{MPIW}$. The intuition behind
the fraction $\frac{n}{\alpha (1 - \alpha)}$ is that it should reflect the
confidence of $\mathcal{L}_{PICP}$ with respect to $n$ and $\alpha$.

\subsubsection{Aggregation Method}
\label{sec:original-aggregation-method}

Given an ensemble of $m$ NN models fitted with respect to $\mathcal{L}_{QD}$
(\ref{eq:original-qde}), we acquire prediction intervals $[\hat{y}_{L}^{(ij)},
\hat{y}_{U}^{(ij)}]$ for sample $i \in \{ 1 \isep n \}$ and model $j \in \{1
\isep m\}$. The following calculations are taken to acquire the final predictions
intervals $[\tilde{y}_{L}^{(i)}, \tilde{y}_{U}^{(i)}]$ that should account also
for the epistemic uncertainty.
\begin{equation*}
  \label{ref:mu-l}
  \begin{aligned}
    \mu_{L}^{(i)}=& \frac{1}{m} \sum_{j=1}^{m} \hat{y}_{L}^{(ij)},
  \end{aligned}
\end{equation*}
\begin{equation*}
  \label{ref:mu-u}
  \begin{aligned}
    \mu_{U}^{(i)}=& \frac{1}{m} \sum_{j=1}^{m} \hat{y}_{U}^{(ij)},
  \end{aligned}
\end{equation*}
\begin{equation*}
  \label{ref:sigma-l}
  \begin{aligned}
    {\sigma_{L}^{(i)}}^{2}=\frac{1}{m-1} \sum_{j=1}^{m}\left(\hat{y}_{L}^{(ij)} - \mu_{L}^{(i)}\right)^{2}
  \end{aligned}
\end{equation*}
\begin{equation*}
  \label{ref:sigma-u}
  \begin{aligned}
    {\sigma_{U}^{(i)}}^{2}=\frac{1}{m-1} \sum_{j=1}^{m}\left(\hat{y}_{U}^{(ij)} - \mu_{U}^{(i)}\right)^{2}
  \end{aligned}
\end{equation*}

These quantities are then combined to generate the final PIs.
In the paper~\cite{pearce:2018}, the aggregation is done as follows:
\begin{equation}
  \label{eq:original-aggregation-method-paper-l}
  \begin{aligned}
    \tilde{y}_{L}^{(i)}=\mu_{L}^{(i)} - 1.96 \cdot \sigma_{L}^{(i)},
  \end{aligned}
\end{equation}
\begin{equation}
  \label{eq:original-aggregation-method-paper-u}
  \begin{aligned}
    \tilde{y}_{U}^{(i)}=\mu_{U}^{(i)} + 1.96 \cdot \sigma_{U}^{(i)}.
  \end{aligned}
\end{equation}

The actual implementation differs in using a standard error of the mean (SEM)
$\sigma_{\bar{y}_{L}}^{(i)}$ (\ref{eq:original-aggregation-method-impl-l}) and
$\sigma_{\bar{y}_{U}}^{(i)}$ (\ref{eq:original-aggregation-method-impl-u})
instead of in the paper presented standard deviation $\sigma_{L}^{(i)}$
(\ref{eq:original-aggregation-method-paper-l}) and $\sigma_{U}^{(i)}$
(\ref{eq:original-aggregation-method-paper-u}) respectively; notice the introduction of the scalar $1/\sqrt{m}$ in both equations.
\begin{equation}
  \label{eq:original-aggregation-method-impl-l}
  \begin{aligned}
    \tilde{y}_{L}^{(i)}=\mu_{L}^{(i)} - 1.96 \cdot \sigma_{L}^{(i)} \cdot \mathbf{\frac{1}{\sqrt{m}}}
  \end{aligned}
\end{equation}
\begin{equation}
  \label{eq:original-aggregation-method-impl-u}
  \begin{aligned}
    \tilde{y}_{U}^{(i)}=\mu_{U}^{(i)} + 1.96 \cdot \sigma_{U}^{(i)} \cdot \mathbf{\frac{1}{\sqrt{m}}}
  \end{aligned}
\end{equation}

Aggregating PIs using the latter equations
(\ref{eq:original-aggregation-method-impl-l}) and
(\ref{eq:original-aggregation-method-impl-u}) results in narrower PIs. Both
aggregation methods lack any theoretical justification:
The lower and upper PI boundaries are aggregated independently, i.e. without mutual consideration, and we consider this as a flaw of the method.
In fact, under a simple assumption of normal posterior distribution that is correctly
captured by $m$ NN models but with shifted PIs with correct $\gamma$ coverage
probability (PICP), it can be shown that both original aggregation methods in
the above equations yield PIs resulting in PICP greater than $\gamma$, and thus
greater MPIW. In the results, the latter aggregation function (eq.~\ref{eq:original-aggregation-method-impl-l} and \ref{eq:original-aggregation-method-impl-u}) was evaluated (annotated as SEM).

\section{METHOD}
\label{sec:method}

In this section, we describe our method annotated as SNM-QD+ that provides a point prediction along with prediction interval as output. First, an ensemble
of neural network models is trained with an extended quality-driven loss
function annotated as QD+ (eq.\ \ref{eq:loss-qd-plus}). Second, the results from each model in the ensemble are aggregated to provide a final result (incorporating the epistemic uncertainty)
by a fitting split normal density functions~\cite{wallis:2014} from each NN's point estimate and PI, and aggregating them into a split normal mixture (annotated as SNM) from which the final PI of coverage probability $\gamma$ is calculated.

\subsection{Quality-Driven Loss Function}
\label{sec:loss-function}

By having a single model for prediction intervals and point estimates, we achieve a
coherency of the results and so avoid the case of two disjoint models learning
different function approximations for prediction intervals and point estimates.
In other words, providing prediction intervals and point estimates as a result
of two disjoint models may not capture the predictive uncertainty of the point
estimate model.

We therefore propose a new loss function
\begin{equation}
  \label{eq:loss-qd-plus}
  \begin{aligned}
  \mathcal{L}_{QD+} ={} & (1 - \lambda_{1}) (1 - \lambda_{2}) \cdot \mathcal{L}_{MPIW} \\
  & + \lambda_{1} (1 - \lambda_{2}) \cdot \mathcal{L}_{PICP} \\
  & + \lambda_{2} \cdot \mathcal{L}_{MSE} \\
  & + \xi \cdot \mathcal{L}_{P}
 \end{aligned}
\end{equation}
where point estimates $\hat{y}^{(i)}$ are optimized by
\begin{equation}
  \label{eq:loss-mse}
  \begin{aligned}
  \mathcal{L}_{MSE} = \frac{1}{n}\sum_{i=1}^{n} \left(\hat{y}^{(i)} - y^{(i)}\right)^{2}
 \end{aligned}
\end{equation}
and the penalty function
\begin{equation}
  \label{eq:penalty-function}
  \mathcal{L}_{P} = \frac{1}{n} \sum_{i=1}^{n} \left[\operatorname{max}(0, \hat{y}_{L}^{(i)} - \hat{y}^{(i)}) + \operatorname{max}(0, \hat{y}^{(i)} - \hat{y}_{U}^{(i)})\right]
\end{equation}
is adding a constraint to enforce their integrity.

We retrofit a slightly simplified version of the loss function by
\citeauthor{pearce:2018} \citeyearpar{pearce:2018}. We introduce an auxiliary
loss $\mathcal{L}_{MSE}$ driving the point estimates, the mean squared error
(MSE) in our particular use case. However, our empirical results showed rather
difficult training process (exhibited already by the original loss function
$\mathcal{L}_{QD}$) and issues with the integrity of the generated output, i.e.
interval crossing and point estimates out of PI bounds. Therefore a constraint
violation penalty function $\mathcal{L}_{P}$ was added which mitigated both
issues significantly.

The hyper-parameter $\lambda_{1} \in (0,1)$ controls the mutual influence between losses $\mathcal{L}_{MPIW}$ (eq.\ \ref{eq:loss-mpiw}) and $\mathcal{L}_{PICP}$ (eq.\ \ref{eq:loss-picp}).
The hyper-parameter $\lambda_{2} \in (0,1)$ controls the influence of the $\mathcal{L}_{MSE}$ (eq.\ \ref{eq:loss-mse}) in relation to the aforementioned $\mathcal{L}_{MPIW}$ and $\mathcal{L}_{PICP}$.
The hyper-parameter $\xi$ of the penalty function $\mathcal{L}_{P}$ (eq.\ \ref{eq:penalty-function}) controls the degree of penalization in case the constraint is violated.

\subsection{Split Normal Aggregation Method}
\label{sec:aggregation-method}

The prediction intervals and point estimates retrieved from the ensemble need
to be aggregated into a final prediction interval, capturing both
the aleatoric and epistemic uncertainty, and a final point estimate.

\begin{figure}[t]
  \vskip 0.2in
  \begin{center}
  \centerline{\includegraphics[width=\columnwidth]{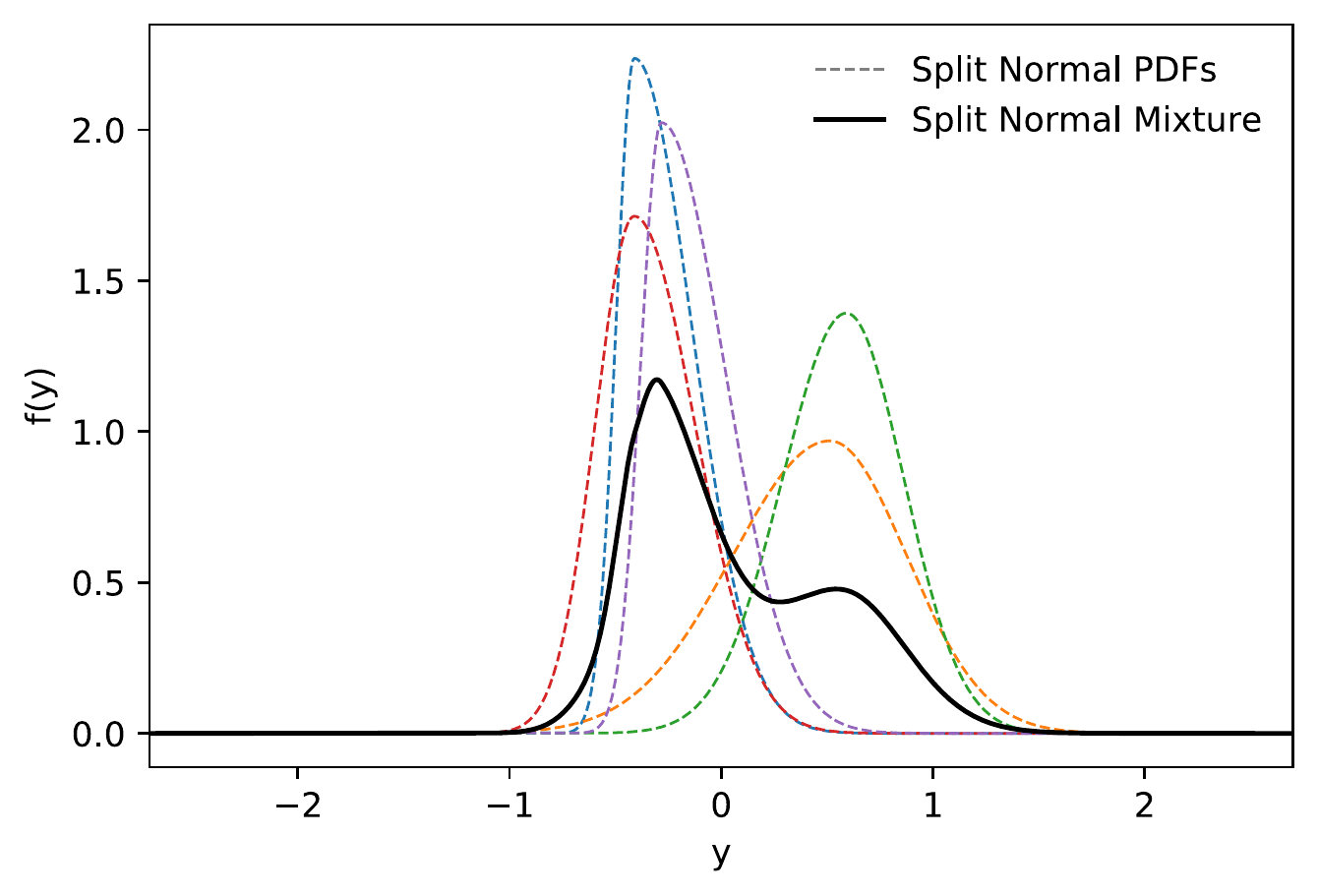}}
  \caption{An example illustrating split normal PDFs (dashed) fitted from an ensemble ($m=5$) of PIs and point estimates, and subsequently aggregated into a split normal mixture (black) from which the final PI is calculated.}
  \label{fig:split-normal-mixture}
  \end{center}
  \vskip -0.2in
\end{figure}

To this end, we assume that the posterior predictive distribution from a single model is a split normal distribution~\cite{wallis:2014}. The split normal distribution or two-piece normal distribution is a result of joining halves of normal distributions with the same mode but different variances. The split normal probability density function (PDF) is defined as
\begin{equation*}
  f_{\mathcal{SN}}\left(x ; \mu, \sigma_{1}, \sigma_{2}\right)=
  \begin{cases}
  A \exp \left(-\frac{(x-\mu)^{2}}{2 \sigma_{1}^{2}}\right) &
  \text{if } x < \mu, \\
  A \exp \left(-\frac{(x-\mu)^{2}}{2 \sigma_{2}^{2}}\right) &
  \text{otherwise},
\end{cases}
\end{equation*}
where $ A=\sqrt{2 / \pi}\left(\sigma_{1}+\sigma_{2}\right)^{-1}$.

Given the lower and upper PI bounds $[\hat{y}_{L}^{(ij)}, \hat{y}_{U}^{(ij)}]$ and the point estimates $\hat{y}^{(ij)}$ from an ensemble, we can fit parameters of a split normal distribution for each ensemble result by optimizing the loss
\begin{equation}
  \label{eq:split-normal-loss}
  \begin{aligned}
  \mathcal{L}_{\mathcal{SN}} = {} & [F_{\mathcal{SN}}(\hat{y}_{L}^{(ij)}; \cdot) - \alpha / 2]^2 \\
  & + [F_{\mathcal{SN}}(\hat{y}_{U}^{(ij)}; \cdot) - (1 - \alpha / 2)]^2
  \end{aligned}
\end{equation}
where the split normal cumulative density function (CDF) is defined as
\begin{equation*}
  F_{\mathcal{SN}}\left(x ; \mu, \sigma_{1}, \sigma_{2}\right) =
  \begin{cases}
  \frac{\sigma_{1} + \operatorname{erf}\left(\frac{x - \mu}{\sqrt{2} \sigma_{1}}\right) \sigma_{1}}{\sigma_{1} + \sigma_{2}} &
  \text{if } x < \mu, \\
  \frac{\sigma_{1} + \operatorname{erf}\left(\frac{x - \mu}{\sqrt{2} \sigma_{2}}\right) \sigma_{2}}{\sigma_{1} + \sigma_{2}} &
  \text{otherwise}.
\end{cases}
\end{equation*}

Through a gradient descent (GD) optimization of eq. (\ref{eq:split-normal-loss}) we estimate the parameters $\sigma^{(ij)}_{1}$ and $\sigma^{(ij)}_{2}$ with respect to $[\hat{y}_{L}^{(ij)}, \hat{y}_{U}^{(ij)}]$ (variable $x$), $\hat{y}^{(ij)}$ (variable $\mu$) and $\alpha$.

When fitting the parameters $\sigma^{(ij)}_{1}$ and $\sigma^{(ij)}_{2}$ with eq.\ (\ref{eq:split-normal-loss}) as the objective, initialization of $(\sigma^{(ij)}_{1}, \sigma^{(ij)}_{2})$ plays an important role for finding the optimal solution. A random initialization does not always yield the optimal solution. Therefore, following the
hypothesis that the parameters of a split normal distribution are close to those of normal distribution, we initialize the $\sigma^{(ij)}_{1}$ and $\sigma^{(ij)}_{2}$ as follows:
\begin{equation*}
  \label{eq:sigma_1-initialization}
  \sigma^{(ij)}_{1} = \frac{\hat{y}_{L}^{(ij)} - \hat{y}^{(ij)}}{\sqrt{2} \operatorname{erf}^{-1}(2 p_{L}-1)},
\end{equation*}
\begin{equation*}
  \label{eq:sigma_2-initialization}
  \sigma^{(ij)}_{2} = \frac{\hat{y}_{U}^{(ij)} - \hat{y}^{(ij)}}{\sqrt{2} \operatorname{erf}^{-1}(2 p_{U}-1)},
\end{equation*}
where $p_{L} = \alpha / 2$ and $p_{U} = 1 - \alpha / 2$. The above is derived from the inverse CDF (quantile function) of a normal distribution:
\begin{equation*}
    F_{\mathcal{N}}^{-1}(p) = \mu + \sigma \sqrt{2} \operatorname{erf}^{-1}(2p - 1)
\end{equation*}
with $p \in (0,1)$.

In our experiments, this has been proven as superior compared to random
initialization.

As a result of the above steps, we acquire an ensemble of $m$ split normal PDFs $\left\{f_{\mathcal{SN}}\left(x; .\right)\right\}_{j=1}^{m}$ (for a single sample $i$) from which we create a mixture PDF
\begin{equation*}
    \label{eq:split-normal-mixture}
    f^{(i)}(x)=\frac{1}{m} \sum_{j=1}^{m} f_{\mathcal{SN}}\left( x; \theta^{(ij)} \right).
\end{equation*}
See Figure \ref{fig:split-normal-mixture} for illustration. The final PIs $[\tilde{y}_{L}^{(i)}, \tilde{y}_{U}^{(i)}]$ are calculated from the mixture distribution $f^{(i)}(x)$ by numerically solving
\begin{equation*}
    \label{eq:final-prediction-intervals}
    \tilde{y}_{L}^{(i)} = {F^{-1}}^{(i)}(\alpha / 2) \quad \text{ and } \quad \tilde{y}_{U}^{(i)} = {F^{-1}}^{(i)}(1 - \alpha / 2)
\end{equation*}
where ${F^{-1}}^{(i)}(p)$ is the inverse CDF of the mixture $f^{(i)}(x)$.

The final point estimates are calculated as an equally weighted combination (mean) of point estimates from an ensemble.
\begin{equation*}
  \tilde{y}^{(i)} = \frac{1}{m} \sum_{j=1}^{m} \hat{y}^{(ij)}
\end{equation*}

\section{EXPERIMENTS}
\label{sec:experiments}

\begin{table*}[t]
  \scriptsize
  \centering
  \caption{Ultimately, PIs are compared between SNM-QD+, SEM-QD and MVE, and point estimates are compared between SNM-QD+, MSE and MVE in Table \ref{tab:results-pe}. Additionally, aggregation methods are compared in SNM-QD+ vs. SEM-QD+. All methods are ensembles of 2-layer NNs. The values stand for mean $\pm$ standard error of the mean. The best results (with standard error of the mean taken in consideration) are shown in bold.}
  \label{tab:results-pi}
  \vskip 1.em
  \setlength\tabcolsep{5pt}  
  \begin{tabular}{l|*4c|*4c}
    \toprule
    \multirow{2}{*}{\textbf{Dataset}}
    & \multicolumn{4}{c|}{\textbf{PICP}}
    & \multicolumn{4}{c}{\textbf{MPIW}} \\
    & SNM-QD+ & SEM-QD+ & SEM-QD & MVE
    & SNM-QD+ & SEM-QD+ & SEM-QD & MVE \\
    \midrule
    Boston
    & $0.95\pm0.01$ & $0.97\pm0.01$ & $\mathbf{0.95\pm0.01}$ & $0.88\pm0.01$
    & $1.58\pm0.06$ & $1.82\pm0.07$ & $\mathbf{1.52\pm0.06}$ & $0.89\pm0.01$ \\
    Concrete
    & $\mathbf{0.94\pm0.01}$ & $0.96\pm0.01$ & $0.97\pm0.00$ & $0.96\pm0.00$
    & $\mathbf{0.99\pm0.04}$ & $1.14\pm0.05$ & $1.36\pm0.02$ & $1.14\pm0.02$ \\
    Energy
    & $0.99\pm0.00$ & $0.99\pm0.00$ & $0.99\pm0.01$ & $\mathbf{0.98\pm0.01}$
    & $0.29\pm0.01$ & $0.33\pm0.02$ & $0.48\pm0.03$ & $\mathbf{0.16\pm0.00}$ \\
    Kin8nm
    & $0.97\pm0.00$ & $0.98\pm0.00$ & $0.99\pm0.00$ & $\mathbf{0.97\pm0.00}$
    & $1.07\pm0.01$ & $1.21\pm0.01$ & $1.29\pm0.01$ & $\mathbf{1.04\pm0.01}$ \\
    Naval
    & $1.00\pm0.00$ & $1.00\pm0.00$ & $0.99\pm0.00$ & $\mathbf{1.00\pm0.00}$
    & $0.09\pm0.00$ & $0.10\pm0.00$ & $0.36\pm0.02$ & $\mathbf{0.05\pm0.00}$ \\
    Power
    & $\mathbf{0.95\pm0.00}$ & $0.96\pm0.00$ & $0.96\pm0.00$ & $0.96\pm0.00$
    & $\mathbf{0.80\pm0.00}$ & $0.85\pm0.00$ & $0.87\pm0.00$ & $0.87\pm0.01$ \\
    Protein
    & $\mathbf{0.95\pm0.00}$ & $0.97\pm0.00$ & $0.95\pm0.00$ & $0.98\pm0.00$
    & $\mathbf{2.12\pm0.01}$ & $2.31\pm0.01$ & $2.24\pm0.01$ & $2.82\pm0.10$ \\
    Wine
    & $\mathbf{0.94\pm0.01}$ & $0.95\pm0.01$ & $0.92\pm0.01$ & $0.94\pm0.00$
    & $\mathbf{2.62\pm0.06}$ & $2.92\pm0.07$ & $2.06\pm0.03$ & $2.94\pm0.02$ \\
    Yacht
    & $\mathbf{0.94\pm0.01}$ & $0.96\pm0.01$ & $1.00\pm0.00$ & $0.99\pm0.00$
    & $\mathbf{0.12\pm0.00}$ & $0.13\pm0.00$ & $0.25\pm0.02$ & $0.40\pm0.04$ \\
    Year
    & $0.94\pm\text{NA}$ & $0.96\pm\text{NA}$ & $\mathbf{0.95\pm\text{\textbf{NA}}}$ & $0.96\pm\text{NA}$
    & $2.34\pm\text{NA}$ & $2.54\pm\text{NA}$ & $\mathbf{2.29\pm\text{\textbf{NA}}}$ & $2.75\pm\text{NA}$ \\
    \bottomrule
  \end{tabular}
\end{table*}

\begin{table}[t]
  \scriptsize
  \centering
  \caption{Extension of Table \ref{tab:results-pi} evaluating point estimates of SNM-QD+, MSE and MVE.}
  \label{tab:results-pe}
  \vskip 1.em
  \setlength\tabcolsep{5pt}  
  \begin{tabular}{l|*3c}
    \toprule
    \multirow{2}{*}{\textbf{Dataset}}
    & \multicolumn{3}{c}{\textbf{MSE}} \\
    & MSE & SNM-QD+ & MVE \\
    \midrule
    Boston
    & $\mathbf{0.112\pm0.013}$ & $0.115\pm0.013$ & $0.127\pm0.019$ \\
    Concrete
    & $0.056\pm0.003$ & $\mathbf{0.053\pm0.003}$ & $0.077\pm0.004$ \\
    Energy
    & $0.001\pm0.000$ & $\mathbf{0.001\pm0.000}$ & $0.001\pm0.000$ \\
    Kin8nm
    & $0.060\pm0.001$ & $\mathbf{0.059\pm0.001}$ & $0.062\pm0.001$ \\
    Naval
    & $\mathbf{4.8\text{e-}5\pm0.000}$ & $1.3\text{e-}4\pm0.000$ & $1.2\text{e-}4\pm0.000$ \\  
    Power
    & $\mathbf{0.042\pm0.001}$ & $0.050\pm0.001$ & $0.048\pm0.001$ \\
    Protein
    & $\mathbf{0.310\pm0.002}$ & $0.361\pm0.004$ & $0.445\pm0.023$ \\
    Wine
    & $\mathbf{0.597\pm0.015}$ & $0.616\pm0.018$ & $0.621\pm0.015$ \\
    Yacht
    & $0.002\pm0.000$ & $\mathbf{0.001\pm0.000}$ & $0.003\pm0.000$ \\
    Year
    & $0.637\pm\text{NA}$ & $\mathbf{0.636\pm\text{\textbf{NA}}}$ & $0.639\pm\text{NA}$ \\
    \bottomrule
  \end{tabular}
\end{table}

Our proposed loss function QD+ (Section~\ref{sec:loss-function}) and aggregation method SNM (Section~\ref{sec:aggregation-method}) are compared with the original QDE~\cite{pearce:2018}, specifically with the loss function QD (Section~\ref{sec:original-loss-function}) and aggregation method SEM (Section~\ref{sec:original-aggregation-method}). We also compare against an ensemble of NNs optimizing the parametrization of a normal distribution~\cite{lakshminarayanan:2017}, referred as a mean-variance estimator (MVE). Additionally, we compare point estimates generated by our method SNM-QD+ with an ensemble of NNs optimizing MSE (eq.\ \ref{eq:loss-mse}) for point estimates solely.

We follow a similar experimental setup initially set by \citeauthor{hernandez:2015}
\yrcite{hernandez:2015} and also followed in related works~\cite{gal:2016,
lakshminarayanan:2017, pearce:2018, tagasovska:2019}.

\subsection{Datasets}
\label{sec:datasets}

Ten open-access benchmark datasets from the UCI dataset repository are used
\cite{dua:2019}. For each dataset we have created 20 shuffled versions across which we have evaluated the methods, i.e. 20 trials. Exceptions are the Year dataset with 1 trial (without shuffling) and the Protein dataset with 5 trials. The datasets are standardized (to zero mean and unit variance) based on the
training set. The input of evaluation measures is standardized based on the full dataset for comparability across different trials.

\subsection{Hyper-Parameters}
\label{sec:hyper-parameters}

We optimize for 95\% prediction intervals, i.e. $\alpha = 0.05$.

To not add any competitive advantage to our method, we keep the NN sizes identical across different models. Note that this may be a disadvantage for our model, given the complexity of the loss function QD+.
This potentially sub-optimal setting should be kept in mind as our comparison's objective is feasibility when compared to models specialized either to generate point estimates or PIs (and not both at the same time).

The ensemble size is $m=5$. The size of mini-batches is $n=100$ ($n=1000$ for the Year dataset). All NNs have 2 hidden layers with 50 units (100 units for Protein and Year datasets) and ReLU activation functions.

For the hyper-parameter search (HPS), we do not follow the legacy of \citeauthor{hernandez:2015} \yrcite{hernandez:2015}. The HPS is performed only on a single concrete shuffled version of a dataset with excluded 10\% test set, i.e. training set remains. Given the training set, the hyper-parameters (HPs) are validated on 5 shuffled 90\% and 10\% splits (i.e. 81\% and 9\% of the complete dataset) for training and validation, respectively. This deviates opposed to the originally suggested single 80\% and 20\% split (i.e. 72\% and 18\% of the complete dataset) without cross-validation. The change of the hyper-parameter search setup was motivated by the difficulty to find good HPs using the original setup, especially in smaller datasets (such as Boston, Concrete, Energy, Wine and Yacht).

A random HPS was performed on the following hyper-parameters (depending on the model): learning rate, decay rate, $\lambda_{1}$, $\lambda_{2}$, epochs. $\xi$ is set to $10$ for all experiments. The remaining settings are consistent with those of \citeauthor{pearce:2018} \yrcite{pearce:2018}.

The applied heuristic for selecting the best hyper-parameters was as follows: If mean PICP is equal to $\gamma \pm 0.01$, then HPs with the lowest MPIW and MSE were selected. Alternatively, also heuristic considering mean and standard deviation of PICP proved good results. If neither of the criteria were fulfilled, the best achieved PICP was selected. Also, the training process on the validation set was analyzed for steady convergence and overfitting.
Note that the loss cannot be used to select HPs because $\lambda_{1}$, $\lambda_{2}$ and $\xi$ scale parts of the loss function; hence the loss is associated with particular values of $\lambda_{1}$, $\lambda_{2}$ and $\xi$.

Due to a smaller number of HPS trials (max. 300), we did not always find well-performing HPs. Therefore, some experiments (QD+ for the datasets Energy, Kin8nm, Naval, Protein and Year) were manually fine-tuned. Adjusting hyper-parameters ($\lambda_1$ and $\lambda_2$, learning rate, decay rate and number of epochs) has proven to be intuitive by analyzing PICP, MPIW and MSE on a validation set. We resorted to manual fine-tuning only if a model was considerably under-performing compared to the original QDE or compared to MVE. There was no attempt to optimize to the model's full potential.

\subsection{Evaluation}
\label{sec:evaluation}

The input of evaluation measures is standardized based on the full dataset for comparability across different trials. The following measures are used to assess and compare models (QD+, QD and MVE) and aggregation methods (SNM and SEM): PICP, MPIW and mean squared error (MSE).

\subsection{Results}
\label{sec:results}

Tables~\ref{tab:results-pi} and \ref{tab:results-pe} show the results of comparing the two models QD+ and QD, combined with the aggregation methods SNM and SEM. We have also included a model that generates only point estimates (denoted MSE), and MVE generating both PIs and point estimates. Consider SNM-QD+ vs. SEM-QD for a comparison between our method and the original QDE method~\cite{pearce:2018}, respectively. SNM-QD+ vs. SEM-QD+ gives a comparison between our split normal aggregation method and the original aggregation method, respectively. SNM-QD+ vs. MVE provides a comparison of methods generating both PIs and point estimates. Finally, Table \ref{tab:results-pe} shows the evaluation of the point estimates between SNM-QD+, MSE and MVE. The best results (with standard error of the mean taken in consideration) are presented using bold font. Supplement~\ref{sec:additional-results} provides the evaluation of non-aggregated PIs.

Overall, the results demonstrate that a complex multi-objective quality-driven loss function (QD+) can deliver well-calibrated PIs when the split normal mixture (SNM) is employed as the aggregation method. The proposed method SNM-QD+ performs best with respect to PIs and competitively with respect to point estimates. We argue that the quality of the point estimates can be improved by increasing the capacity of NNs or by hyper-parameter fine-tuning.

The results clearly show that the split normal aggregation method (SNM) yields well-calibrated PIs, while the original aggregation method (SEM) generally tends to generate wider PIs.
Contrary to SEM, SNM aggregation method does not treat PI boundaries independently. It fits flexible enough (asymmetric) distributions within a versatile mixture distribution that is, as the results show, a sufficient approximation of the posterior predictive distribution.
Note that the SNM aggregation method is not necessarily tied to the QD+ loss, but can be applied in other situations as well.
A potential drawback of SNM is the apparent computational overhead of fitting the split normal mixture when the number of samples is large. However, this learning task is parallelizable and distributable; hence the wall-clock time spent on the task is negligible.

The training process of QD+ is, contrary to the original QD, quite robust.
To give a particular example with the Protein dataset, training 2-layer NNs with the original QDE required approximately one retry per run due to interval crossing or high loss value (stuck in local minimum). No retries were required when using QD+. Details are given in Supplement~\ref{sec:robustness}.
Furthermore, the original QD is sensitive to parameter initialization. Overall, the penalty function $\mathcal{L}_{P}$ (eq.~\ref{eq:penalty-function}) in QD+ results in significant improvements of the stability of the training process and strengthens the integrity of the output (mitigates the undesired interval crossing and point estimates out of the PI bounds).
Supplement~\ref{sec:sensitivity-analysis} provides a sensitivity analysis together with an ablation study clearly demonstrating the importance of the penalty function.

Since we compare our results with \citeauthor{pearce:2018} \yrcite{pearce:2018} and \citeauthor{lakshminarayanan:2017} \yrcite{lakshminarayanan:2017}, our work is also comparable with Bayesian approach by \citeauthor{hernandez:2015} \yrcite{hernandez:2015}, and partly also with \citeauthor{tagasovska:2019} \yrcite{tagasovska:2019}.

\subsection{Implementation}
\label{sec:implementation}

Our implementation is shared\footnote{\url{https://github.com/tarik/pi-snm-qde}} and results fully reproducible. To embrace reproducibility, we use the Python package {Sacred}~\cite{sacred:2017}. PyTorch~\cite{pytorch:2019} is used as the main NN framework and the fitting of a split normal mixture is implemented using JAX~\cite{jax:2018}.

\section{CONCLUSIONS}
\label{sec:conclusions}

The main finding of this paper is that models delivering point estimates together with prediction intervals are competitive to models providing only one or the other.
Motivated by the results of the quality-driven deep ensembles~\cite{pearce:2018} as an alternative to Bayesian methods, we endeavored to address the three key limitations: (1) The inability to generate point estimates; (2) the weakly justified method for aggregation of prediction interval ensembles; (3) the fragile training process.

We propose a new quality-driven loss function generating both prediction intervals and point estimates, and we dramatically increase the robustness of the training process by integrating a penalty function.

A unique and well-founded method fitting a split normal mixture as the aggregate of ensembled neural network output generates well-calibrated prediction intervals accounting for aleatoric and epistemic uncertainty. Moreover, an analytical approach for initializing parameters of a split normal probability density function is proposed that leads to acceleration and dramatically increased success of the fitting process.

With this work, we extend the practitioners toolset for quantifying predictive uncertainty in the regression task.

\section{FUTURE WORK}
\label{sec:future-work}

We end the paper with the following suggestions for future research or possible improvements:

\begin{itemize}
    \item
    We have shown that a single model architecture can be used to generate both point estimates and prediction intervals. We therefore envision a two-step learning process, where one first optimizes the model to generate point estimates and consequently learns to generate prediction intervals. To avoid catastrophic forgetting, it seems appropriate to treat this as a transfer learning problem. The key benefit of this operation is that it will allow us to reuse (pre-trained) NN models that currently only generate point estimates.

    \item
    It seems reasonable to use the NMPIW measure (eq.\ \ref{eq:nmpiw}) in place of MPIW in the definition of the loss function $\mathcal{L}_{MPIW}$ (eq.\ \ref{eq:loss-mpiw}). While it is already relatively intuitive to find HPs manually, and the HPS can be automated efficiently, we hypothesise that using NMPIW may lead to better-scaled hyper-parameters and possibly also to smaller variations across different datasets. The current work uses MPIW to ease the comparison between QD and QD+.

    \item
    The current approach learns to output a reasonable PI that is later used to fit a split normal PDF.
    This could alternatively be simplified so that the output of the neural network was the parameterization of the split normal PDF (or a mixture of split normal PDFs) directly.
    This approach is relevant for use-cases where a full predictive distribution is needed and would require a training objective defined for such situations instead of QD+. Again, the implemented approach was chosen to extend the non-Bayesian branch of research and ease comparison with the work of \citet{pearce:2018}.
\end{itemize}

%

\bibliography{main}

\begin{thebibliography}{24}
\expandafter\ifx\csname natexlab\endcsname\relax\def\natexlab#1{#1}\fi
\expandafter\ifx\csname url\endcsname\relax
  \def\url#1{{\tt #1}}\fi

\bibitem[Graves(2011)]{graves:2011}
Alex Graves.
\newblock {P}ractical {V}ariational {I}nference for {N}eural {N}etworks.
\newblock In {\em Advances in Neural Information Processing Systems 24}, pages
  2348--2356. 2011.

\bibitem[Blundell et~al.(2015)Blundell, Cornebise, Kavukcuoglu, and
  Wierstra]{blundell:2015}
Charles Blundell, Julien Cornebise, Koray Kavukcuoglu, and Daan Wierstra.
\newblock {W}eight {U}ncertainty in {N}eural {N}etwork.
\newblock In {\em Proceedings of the 32nd International Conference on Machine
  Learning}, volume~37 of {\em PMLR}, pages 1613--1622, 2015.

\bibitem[Hernandez-Lobato and Adams(2015)]{hernandez:2015}
Jose~Miguel Hernandez-Lobato and Ryan Adams.
\newblock {P}robabilistic {B}ackpropagation for {S}calable {L}earning of
  {B}ayesian {N}eural {N}etworks.
\newblock In {\em Proceedings of the 32nd International Conference on Machine
  Learning}, volume~37 of {\em PMLR}, pages 1861--1869, 2015.

\bibitem[Krueger et~al.(2017)Krueger, Huang, Islam, Turner, Lacoste, and
  Courville]{krueger:2017}
David Krueger, Chin-Wei Huang, Riashat Islam, Ryan Turner, Alexandre Lacoste,
  and Aaron Courville.
\newblock {B}ayesian {H}ypernetworks, 2017, arXiv:1710.04759.

\bibitem[Louizos and Welling(2017)]{louizos:2017}
Christos Louizos and Max Welling.
\newblock {M}ultiplicative {N}ormalizing {F}lows for {V}ariational {B}ayesian
  {N}eural {N}etworks.
\newblock In {\em Proceedings of the 34th International Conference on Machine
  Learning}, volume~70 of {\em PMLR}, pages 2218--2227, 2017.

\bibitem[Pawlowski et~al.(2017)Pawlowski, Brock, Lee, Rajchl, and
  Glocker]{pawlowski:2017}
Nick Pawlowski, Andrew Brock, Matthew C.~H. Lee, Martin Rajchl, and Ben
  Glocker.
\newblock {I}mplicit {W}eight {U}ncertainty in {N}eural {N}etworks, 2017,
  arXiv:1711.01297.

\bibitem[Wu et~al.(2019)Wu, Nowozin, Meeds, Turner, Hernandez-Lobato, and
  Gaunt]{wu:2019}
Anqi Wu, Sebastian Nowozin, Edward Meeds, Richard~E. Turner, Jose~Miguel
  Hernandez-Lobato, and Alexander~L. Gaunt.
\newblock {D}eterministic {V}ariational {I}nference for {R}obust {B}ayesian
  {N}eural {N}etworks.
\newblock In {\em International Conference on Learning Representations}, 2019.

\bibitem[Izmailov et~al.(2019)Izmailov, Maddox, Kirichenko, Garipov, Vetrov,
  and Wilson]{izmailov:2019}
Pavel Izmailov, Wesley Maddox, Polina Kirichenko, Timur Garipov, Dmitry Vetrov,
  and Andrew~Gordon Wilson.
\newblock {S}ubspace {I}nference for {B}ayesian {D}eep {L}earning.
\newblock In {\em Proceedings of the 35th Conference on Uncertainty in
  Artificial Intelligence (UAI)}, 2019.

\bibitem[Yao et~al.(2019)Yao, Pan, Ghosh, and Doshi-Velez]{yao:2019}
Jiayu Yao, Weiwei Pan, Soumya Ghosh, and Finale Doshi-Velez.
\newblock {Q}uality of {U}ncertainty {Q}uantification for {B}ayesian {N}eural
  {N}etwork {I}nference, 2019, arXiv:1906.09686.

\bibitem[Lakshminarayanan et~al.(2017)Lakshminarayanan, Pritzel, and
  Blundell]{lakshminarayanan:2017}
Balaji Lakshminarayanan, Alexander Pritzel, and Charles Blundell.
\newblock {S}imple and {S}calable {P}redictive {U}ncertainty {E}stimation using
  {D}eep {E}nsembles.
\newblock In {\em Advances in Neural Information Processing Systems 30}, pages
  6402--6413. 2017.

\bibitem[Pearce et~al.(2018)Pearce, Brintrup, Zaki, and Neely]{pearce:2018}
Tim Pearce, Alexandra Brintrup, Mohamed Zaki, and Andy Neely.
\newblock {H}igh-{Q}uality {P}rediction {I}ntervals for {D}eep {L}earning: {A}
  {D}istribution-{F}ree, {E}nsembled {A}pproach.
\newblock In {\em Proceedings of the 35th International Conference on Machine
  Learning}, volume~80 of {\em PMLR}, pages 4075--4084, 2018.

\bibitem[Tagasovska and Lopez-Paz(2019)]{tagasovska:2019}
Natasa Tagasovska and David Lopez-Paz.
\newblock {S}ingle-{M}odel {U}ncertainties for {D}eep {L}earning.
\newblock In {\em Advances in Neural Information Processing Systems 32}, pages
  6414--6425. 2019.

\bibitem[Khosravi et~al.(2011)Khosravi, Nahavandi, Creighton, and
  Atiya]{khosravi:2011}
A.~Khosravi, S.~Nahavandi, D.~Creighton, and A.~F. Atiya.
\newblock {L}ower {U}pper {B}ound {E}stimation {M}ethod for {C}onstruction of
  {N}eural {N}etwork-{B}ased {P}rediction {I}ntervals.
\newblock {\em IEEE Transactions on Neural Networks}, 22\penalty0 (3):\penalty0
  337--346, 2011.

\bibitem[Wallis(2014)]{wallis:2014}
Kenneth~F. Wallis.
\newblock {T}he {T}wo-{P}iece {N}ormal, {B}inormal, or {D}ouble {G}aussian
  {D}istribution: {I}ts {O}rigin and {R}ediscoveries.
\newblock {\em Statistical Science}, 29\penalty0 (1):\penalty0 106--112, 2014.

\bibitem[Kiureghian and Ditlevsen(2009)]{kiureghian:2009}
Armen~Der Kiureghian and Ove Ditlevsen.
\newblock {A}leatory or epistemic? {D}oes it matter?
\newblock {\em Structural Safety}, 31\penalty0 (2):\penalty0 105--112, 2009.

\bibitem[Malinin and Gales(2018)]{malinin:2018}
Andrey Malinin and Mark Gales.
\newblock {P}redictive {U}ncertainty {E}stimation via {P}rior {N}etworks.
\newblock In {\em Advances in Neural Information Processing Systems 31}, pages
  7047--7058. 2018.

\bibitem[Gal and Ghahramani(2016)]{gal:2016}
Yarin Gal and Zoubin Ghahramani.
\newblock {D}ropout as a {B}ayesian {A}pproximation: {R}epresenting {M}odel
  {U}ncertainty in {D}eep {L}earning.
\newblock In {\em Proceedings of The 33rd International Conference on Machine
  Learning}, volume~48 of {\em PMLR}, pages 1050--1059, 2016.

\bibitem[Fort et~al.(2019)Fort, Hu, and Lakshminarayanan]{fort:2019}
Stanislav Fort, Huiyi Hu, and Balaji Lakshminarayanan.
\newblock {D}eep {E}nsembles: {A} {L}oss {L}andscape {P}erspective, 2019,
  arXiv:1912.02757.

\bibitem[Ovadia et~al.(2019)Ovadia, Fertig, Ren, Nado, Sculley, Nowozin,
  Dillon, Lakshminarayanan, and Snoek]{ovadia:2019}
Yaniv Ovadia, Emily Fertig, Jie Ren, Zachary Nado, D~Sculley, Sebastian
  Nowozin, Joshua~V. Dillon, Balaji Lakshminarayanan, and Jasper Snoek.
\newblock Can you trust your model\textquotesingle s uncertainty? {E}valuating
  predictive uncertainty under dataset shift.
\newblock In {\em Advances in Neural Information Processing Systems 32}, pages
  13991--14002. 2019.

\bibitem[Koenker(2005)]{koenker:2005}
Roger Koenker.
\newblock {\em {Q}uantile {R}egression}.
\newblock Econometric Society Monographs. Cambridge University Press, 2005.

\bibitem[Dua and Graff(2017)]{dua:2019}
Dheeru Dua and Casey Graff.
\newblock {UCI} {M}achine {L}earning {R}epository, 2017.

\bibitem[Greff et~al.(2017)Greff, Klein, Chovanec, Hutter, and
  Schmidhuber]{sacred:2017}
Klaus Greff, Aaron Klein, Martin Chovanec, Frank Hutter, and J\"urgen
  Schmidhuber.
\newblock {T}he {S}acred {I}nfrastructure for {C}omputational {R}esearch.
\newblock In {\em {P}roceedings of the 16th {P}ython in {S}cience
  {C}onference}, pages 49--56, 2017.

\bibitem[Paszke et~al.(2019)Paszke, Gross, Massa, Lerer, Bradbury, Chanan,
  Killeen, Lin, Gimelshein, Antiga, Desmaison, Kopf, Yang, DeVito, Raison,
  Tejani, Chilamkurthy, Steiner, Fang, Bai, and Chintala]{pytorch:2019}
Adam Paszke, Sam Gross, Francisco Massa, Adam Lerer, James Bradbury, Gregory
  Chanan, Trevor Killeen, Zeming Lin, Natalia Gimelshein, Luca Antiga, Alban
  Desmaison, Andreas Kopf, Edward Yang, Zachary DeVito, Martin Raison, Alykhan
  Tejani, Sasank Chilamkurthy, Benoit Steiner, Lu~Fang, Junjie Bai, and Soumith
  Chintala.
\newblock {PyTorch}: {A}n {I}mperative {S}tyle, {H}igh-{P}erformance {D}eep
  {L}earning {L}ibrary.
\newblock In {\em Advances in Neural Information Processing Systems 32}, pages
  8026--8037. 2019.

\bibitem[Bradbury et~al.(2018)Bradbury, Frostig, Hawkins, Johnson, Leary,
  Maclaurin, and Wanderman-Milne]{jax:2018}
James Bradbury, Roy Frostig, Peter Hawkins, Matthew~James Johnson, Chris Leary,
  Dougal Maclaurin, and Skye Wanderman-Milne.
\newblock {JAX}: {C}omposable {T}ransformations of {P}ython+{N}um{P}y
  {P}rograms, 2018.
\newblock URL \url{http://github.com/google/jax}.

\end{thebibliography}
\bibliographystyle{hunsrtnat}

\setcounter{figure}{0}
\renewcommand{\thefigure}{\Alph{figure}}
\setcounter{table}{0}
\renewcommand{\thetable}{\Alph{table}}

\begin{appendices}
\appendixpage

\section{Additional Results}
\label{sec:additional-results}

Related to Table \ref{tab:results-pi}, Table \ref{tab:results-non-aggregated}
shows the evaluation of non-aggregated PIs (not accounting for epistemic
uncertainty) for the QD+ and QD models.

\begin{table}[h]
  \scriptsize
  \centering
  \caption{Evaluation of non-aggregated PIs (2-layer NNs).}
  \label{tab:results-non-aggregated}
  \vskip 1.em
  \setlength\tabcolsep{4pt}  
  \begin{tabular}{l|*2c|*2c}
    \toprule
    \multirow{2}{*}{\textbf{Dataset}}
    & \multicolumn{2}{c|}{\textbf{PICP}}
    & \multicolumn{2}{c}{\textbf{MPIW}} \\
    & QD+ & QD
    & QD+ & QD \\
    \midrule
    Boston
    & $0.85\pm0.01$ & $0.88\pm0.01$
    & $1.22\pm0.04$ & $1.12\pm0.02$ \\
    Concrete
    & $0.86\pm0.01$ & $0.89\pm0.00$
    & $0.79\pm0.03$ & $1.03\pm0.01$ \\
    Energy
    & $0.90\pm0.01$ & $0.93\pm0.00$
    & $0.21\pm0.01$ & $0.27\pm0.01$ \\
    Kin8nm
    & $0.91\pm0.00$ & $0.92\pm0.00$
    & $0.95\pm0.01$ & $1.01\pm0.01$ \\
    Naval
    & $0.98\pm0.00$ & $0.94\pm0.00$
    & $0.08\pm0.00$ & $0.25\pm0.00$ \\
    Power
    & $0.94\pm0.00$ & $0.93\pm0.00$
    & $0.79\pm0.00$ & $0.75\pm0.00$ \\
    Protein
    & $0.93\pm0.00$ & $0.93\pm0.00$
    & $2.03\pm0.01$ & $2.11\pm0.01$ \\
    Wine
    & $0.91\pm0.00$ & $0.91\pm0.00$
    & $2.31\pm0.03$ & $1.81\pm0.01$ \\
    Yacht
    & $0.76\pm0.01$ & $0.89\pm0.01$
    & $0.07\pm0.00$ & $0.15\pm0.01$ \\
    Year
    & $0.93\pm0.00$ & $0.93\pm0.00$
    & $2.27\pm0.00$ & $2.07\pm0.02$ \\
    \bottomrule
  \end{tabular}
  \vskip 1.em
\end{table}

In Table \ref{tab:qde-results}, we show the reproduced results of the original
QDE method following the exact experimental setting as implemented in
\citeauthor{pearce:2018} \yrcite{pearce:2018}. These results are included to
offer a comparison between the 1-layer and 2-layer models (Table
\ref{tab:qde-results} and Table \ref{tab:results-pi}, respectively). It also
serves as a verification of our re-implementation of the method by
\citet{pearce:2018}.

\begin{table}[h]
  \scriptsize
  \centering
  \caption{Original QDE with 1-layer NNs as a reference.}
  \label{tab:qde-results}
  \vskip 1.em
  \setlength\tabcolsep{4pt}  
  \begin{tabular}{l|*2c|*2c}
    \toprule
    \multirow{2}{*}{\textbf{Dataset}}
    & \multicolumn{2}{c|}{\textbf{PICP}}
    & \multicolumn{2}{c}{\textbf{MPIW}} \\
    & SEM-QD & QD
    & SEM-QD & QD \\
    \midrule
    Boston
    & $0.90\pm0.01$ & $0.80\pm0.01$
    & $1.01\pm0.01$ & $0.81\pm0.01$ \\
    Concrete
    & $0.92\pm0.01$ & $0.84\pm0.00$
    & $1.01\pm0.01$ & $0.83\pm0.00$ \\
    Energy
    & $0.97\pm0.00$ & $0.91\pm0.00$
    & $0.45\pm0.01$ & $0.38\pm0.01$ \\
    Kin8nm
    & $0.96\pm0.00$ & $0.89\pm0.00$
    & $1.24\pm0.00$ & $1.02\pm0.00$ \\
    Naval
    & $0.99\pm0.00$ & $0.96\pm0.00$
    & $0.25\pm0.02$ & $0.17\pm0.00$ \\
    Power
    & $0.95\pm0.00$ & $0.94\pm0.00$
    & $0.85\pm0.00$ & $0.81\pm0.00$ \\
    Protein
    & $0.95\pm0.00$ & $0.94\pm0.00$
    & $2.26\pm0.00$ & $2.17\pm0.00$ \\
    Wine
    & $0.93\pm0.01$ & $0.91\pm0.00$
    & $2.31\pm0.02$ & $2.04\pm0.01$ \\
    Yacht
    & $0.95\pm0.01$ & $0.86\pm0.01$
    & $0.15\pm0.00$ & $0.10\pm0.00$ \\
    Year
    & $0.95\pm\text{NA}$ & $0.93\pm0.00$
    & $2.41\pm\text{NA}$ & $2.22\pm0.01$ \\
    \bottomrule
  \end{tabular}
\end{table}

Table \ref{tab:dataset-and-sem-qd-pe} considers a simple baseline with point estimates constructed as the mean of PIs from SEM-QD (annotated SEM-QD*). This simple baseline is underperforming compared to the results in Table \ref{tab:results-pe}.

\begin{table}[h]
  \scriptsize
  \centering
  \caption{Extension of Table \ref{tab:results-pe} provides a simple baseline with point estimates constructed as the mean of PIs from SEM-QD (annotated SEM-QD*). Also, the sizes $|\mathcal{D}|$ and input dimensions $d$ of the datasets are presented.}
  \label{tab:dataset-and-sem-qd-pe}
  \vskip 1.em
  \setlength\tabcolsep{4pt}  
  \begin{tabular}{l*2c|*1c}
    \toprule
    \multicolumn{3}{c|}{\textbf{Dataset}}
    & \multicolumn{1}{c}{\textbf{MSE}} \\
    Name
    & $|\mathcal{D}|$ & $d$
    & SEM-QD* \\
    \midrule
    Boston
    & $506$ & $13$
    & $0.209\pm0.030$ \\
    Concrete
    & $1030$ & $8$
    & $0.102\pm0.005$ \\
    Energy
    & $768$ & $8$
    & $0.028\pm0.008$ \\
    Kin8nm
    & $8192$ & $8$
    & $0.067\pm0.001$ \\
    Naval
    & $11934$ & $16$
    & $0.012\pm0.001$ \\
    Power
    & $9568$ & $4$
    & $0.054\pm0.001$ \\
    Protein
    & $45730$ & $9$
    & $0.666\pm0.004$ \\
    Wine
    & $1599$ & $11$
    & $0.804\pm0.021$ \\
    Yacht
    & $308$ & $6$
    & $0.003\pm0.001$ \\
    Year
    & $515345$ & $90$
    & $0.686\pm\text{NA}$ \\
    \bottomrule
  \end{tabular}
\end{table}

\section{Robustness}
\label{sec:robustness}

The robustness of training process of QD+ vs. the fragility of training process of QD is shown in Table \ref{tab:robustness}. The contribution of the penalty function $\mathcal{L}_{P}$ to the robustness is shown in Figure \ref{fig:sensitivity-analysis} of the following section.

\begin{table}[h]
  \scriptsize
  \centering
  \caption{Extension of Tables \ref{tab:results-pi} and \ref{tab:results-pe} showing the number of failed/repeated training attempts of QD+ and QD. The targeted model count equals to an ensemble size $m$ times number of trials $t$.}
  \label{tab:robustness}
  \vskip 1.em
  \setlength\tabcolsep{6pt}  
  \begin{tabular}{l|*2c|*1c}
    \toprule
    \multirow{2}{*}{\textbf{Dataset}}
    & \multicolumn{2}{c|}{\textbf{Failures/Retries}}
    & \multicolumn{1}{c}{\textbf{Target Model Count}} \\
    & QD+ & QD
    & $m \cdot t$ \\
    \midrule
    Boston
    & $0$ & $3$
    & $5 \cdot 20$ \\
    Concrete
    & $0$ & $2$
    & $5 \cdot 20$ \\
    Energy
    & $0$ & $22$
    & $5 \cdot 20$ \\
    Kin8nm
    & $0$ & $2$
    & $5 \cdot 20$ \\
    Naval
    & $0$ & $10$
    & $5 \cdot 20$ \\
    Power
    & $0$ & $6$
    & $5 \cdot 20$ \\
    Protein
    & $0$ & $22$
    & $5 \cdot 5$ \\
    Wine
    & $0$ & $29$
    & $5 \cdot 20$ \\
    Yacht
    & $0$ & $33$
    & $5 \cdot 20$ \\
    Year
    & $0$ & $18$
    & $5 \cdot 1$ \\
    \bottomrule
  \end{tabular}
\end{table}

\section{Sensitivity Analysis}
\label{sec:sensitivity-analysis}

\begin{figure*}[p]
  \begin{subfigure}[t]{.5\textwidth}
    \vskip 0.2in
    \begin{center}
    \centerline{\includegraphics[width=.9\columnwidth]{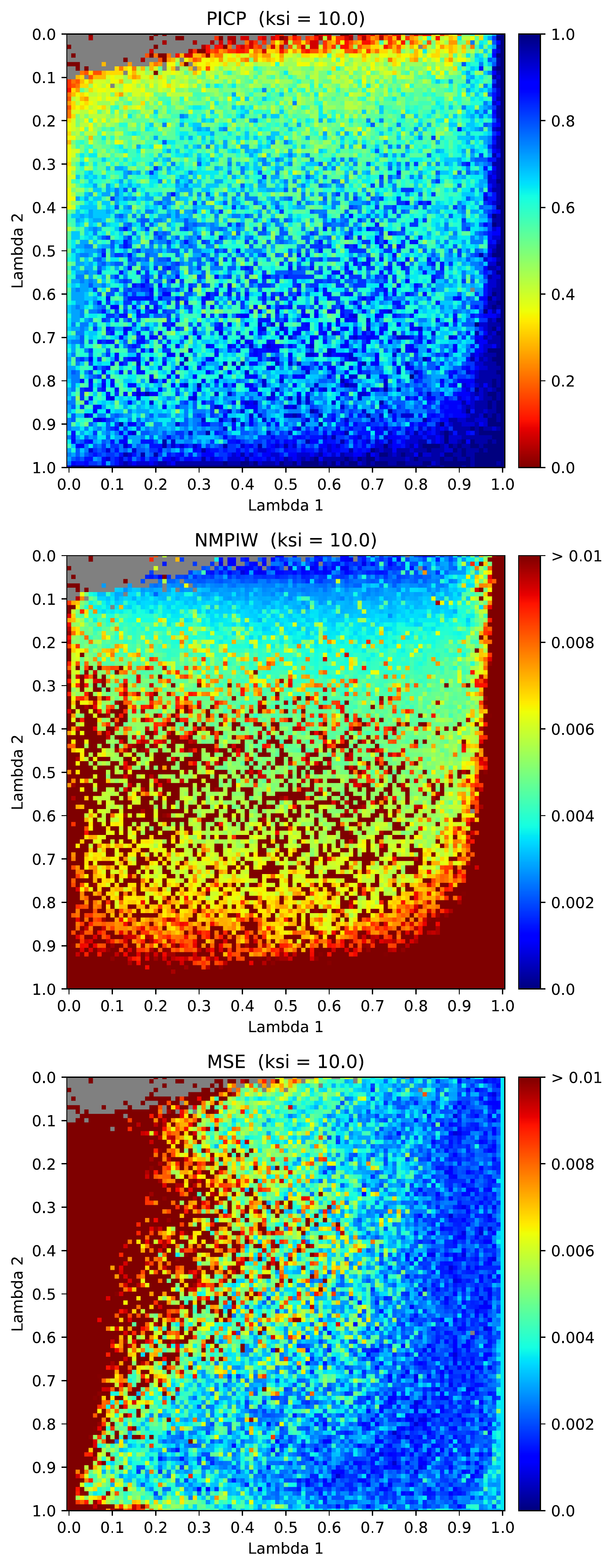}}
    \caption{$\xi = 10$}
    \label{fig:with-penalty-function}
    \end{center}
    \vskip -0.2in
  \end{subfigure}
  \begin{subfigure}[t]{.5\textwidth}
    \vskip 0.2in
    \begin{center}
    \centerline{\includegraphics[width=0.9\columnwidth]{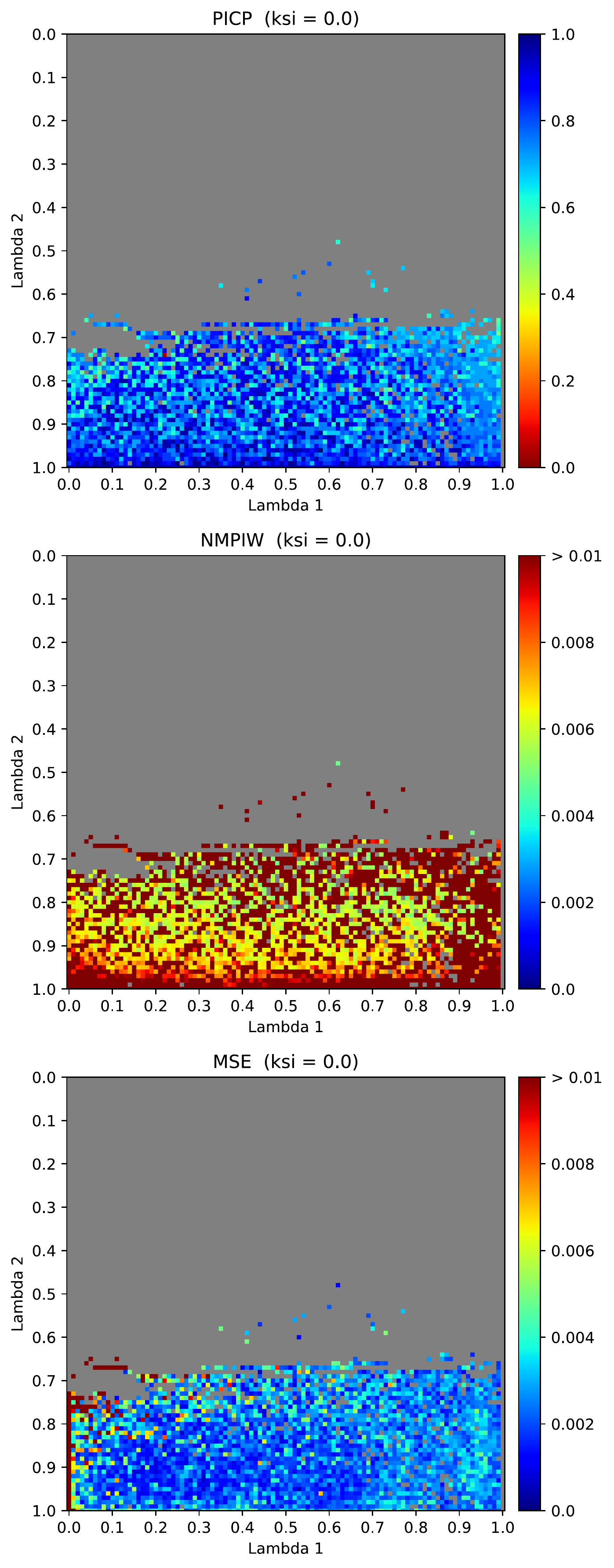}}
    \caption{$\xi = 0$}
    \label{fig:without-penalty-function}
    \end{center}
    \vskip -0.2in
  \end{subfigure}
  \caption{Sensitivity analysis of QD+ on Yacht dataset with respect to hyper-parameters $\lambda_1$ and $\lambda_2$ with (\subref{fig:with-penalty-function}) and without (\subref{fig:without-penalty-function}) an active penalty function $\mathcal{L}_{P}$. Gray color stands for a failed training process, i.e. high loss or outputs violating the semantic integrity (PI crossing or point estimates
outside the PI bounds).}
  \label{fig:sensitivity-analysis}
\end{figure*}

Figure \ref{fig:sensitivity-analysis} provides a sensitivity analysis with respect to hyper-parameters $\lambda_1$, $\lambda_2$ and $\xi$. It visualizes the measures PICP, NMPIW and MSE in relation to changing hyper-parameters of the QD+ model. The analysis was performed on the Yacht dataset.
The sensitivity analysis is also an ablation study since the border values, i.e. $\lambda_1, \lambda_2 \in \{0, 1\}$ and $\xi = 0$, disable certain parts of the $\mathcal{L}_{QD+}$ loss function (eq. \ref{eq:loss-qd-plus}).

Figure \ref{fig:with-penalty-function} would guide us to choose $\lambda_1$ around $0.975$ and $\lambda_2$ around $0.05$. Leaving out the penalty function from the loss function (i.e. $\xi = 0$) leads to a non-converging training process for $\lambda_2$ below approx. $0.7$ and more noisy measures in the remaining hyper-parameter space (Figure \ref{fig:without-penalty-function}). It demonstrates the critical importance of the penalty function $\mathcal{L}_{P}$ (eq. \ref{eq:penalty-function}), thus, supporting the claims about the robustness together with the evidence in Table \ref{tab:robustness}.

\end{appendices}

\end{document}